\documentclass{ceurart}

\usepackage{amsmath}
\usepackage{amssymb}
\usepackage{amsthm}
\usepackage{mathtools}
\usepackage{xspace}
\usepackage{multirow}
\usepackage{subcaption}
\usepackage{algorithm}
\usepackage{algorithmic}

\def\wattlayer{{WattLayer}\xspace}
\newcommand{\adrien}[1]{{#1}}

\providecommand{\shortcite}{\cite}

\begin{document}

%% Rights management information (CC BY 4.0 is the CEUR-WS default).
\copyrightyear{2026}
\copyrightclause{Copyright for this paper by its authors.
	Use permitted under Creative Commons License Attribution 4.0
	International (CC BY 4.0).}

%% Workshop / venue information.
\conference{SuRE'26: 1st Workshop on Sustainability and Resource-Efficiency of
	Artificial Intelligence, co-located with IJCAI 2026,
	August 17, 2026, Bremen, Germany}

%% NOTE: the literal word "WattLayer" is used here instead of the \wattlayer
%% macro because the ceurart class injects the title into the PDF/A XMP metadata,
%% where custom macros are not available. The rendered title is identical to the
%% original \title{\wattlayer: ...}.
\title{WattLayer: Get Layers Right to Estimate Inference Energy of Neural Networks}

\author[1,2,3]{Adrien Sardi}[%
	email=adrien.sardi@nokia.com,
    orcid=0009-0004-2226-6253,
    url=https://aupeka.github.io/
]
\author[1]{Marie Line Alberi-Morel}[%
	email=marie\string\_line.alberi-morel@nokia-bell-labs.com,
    orcid=0000-0002-3843-7617,
]
\author[2]{Sara Alouf}[%
	email=sara.alouf@inria.fr,
    orcid=0009-0009-6643-0568,
    url=https://www-sop.inria.fr/members/Sara.Alouf/
]
\author[3]{Frédéric Giroire}[%
	email=frederic.giroire@cnrs.fr,
    orcid=0000-0002-3727-051X,
    url=https://www-sop.inria.fr/members/Frederic.Giroire/
]
\author[3]{Joanna Moulierac}[%
	email=joanna.moulierac@univ-cotedazur.fr,
    orcid=0000-0002-4367-5839,
    url=https://www-sop.inria.fr/members/Joanna.Moulierac/
]

\address[1]{Bell Labs, Nokia Networks France}
\address[2]{Inria, Université Côte d'Azur, France}
\address[3]{Université Côte d’Azur, CNRS, Inria, I3S, France} % pas de "Sophia Antipolis"
%% https://univ-cotedazur.fr/recherche-innovation/services-aux-chercheurs/signature-scientifique-1
%% L’adresse sera réduite à la seule mention du pays. 

%% NOTE: the ceurart class captures the abstract body verbatim (it is written to
%% \jobname.abs for metadata extraction), so the environment must not contain
%% blank lines or %-comments. The commented-out alternative abstracts present in
%% the original sure26.tex are therefore kept here, outside the environment, for
%% reference only:
%%   % The widespread adoption of Artificial Intelligence (AI) ... (draft 1)
%% The active abstract text below is copied verbatim from sure26.tex.
\begin{abstract}
The widespread adoption of Artificial Intelligence (AI) has led to increasing concerns about energy consumption, yet there is a lack of standardized methodologies to accurately estimate AI inference energy consumption, particularly across various tasks and architectures.
In this study, we propose a task independent, layer wise energy estimation model
for AI architectures.
Our model is evaluated on a large dataset of more than 100,000 layers for 295 neural network architectures across 3 widely used tasks and 3 distinct hardware platforms.
Our approach achieves a median error of $19.6\%$, outperforming state-of-the-art methods.
We further show that layer-wise decomposition generalizes to new tasks without complete retraining, by leveraging shared layers across architectures.
It offers tools, insights, and a precise methodology to empower stakeholders in designing energy-efficient AI systems.
\end{abstract}


\maketitle

\section{Introduction}

Artificial Intelligence (AI) has been rapidly adopted across industries due to recent advances in model accuracy~\cite{maslej2025artificialintelligenceindexreport}. This widespread adoption has led to a substantial increase in energy demand, particularly for large-scale models. For instance, Google reported that machine learning workloads accounted for 10\% to 15\% of its total energy consumption between 2019 and 2021~\shortcite{patterson_carbon_2022}.
AI energy consumption can be broadly divided into two phases: training and inference. While most efficiency efforts have historically focused on training due to its high absolute cost compared to a single inference~\cite{luccioni_power_2024}, the large-scale deployment of AI services has shifted this balance. 
At scale, inference—the phase where models are used—can dominate overall energy consumption. Indeed, reports from Google~\shortcite{patterson_carbon_2022} and Meta~\shortcite{wu_sustainable_2022} indicate that inference can account for approximately 60–65\% of total energy use, reaching up to 90\% in cloud deployments such as Amazon Web Services~\shortcite{chateauvieux_optimize_2022}.
As AI systems become increasingly integrated into real-world applications, energy-aware model design is gaining importance among developers. However, despite this growing need, there is still a lack of standardized tools and methodologies to estimate the energy consumption of AI systems without executing the models. While companies such as Google~\shortcite{elsworth_google_2025} and Mistral AI~\shortcite{mistralai_our_2025} have reported energy usage for large language models, the absence of unified estimation frameworks makes comparisons difficult and limits practical use in local or personal deployment scenarios.

This paper introduces \wattlayer: a task-independent layer-wise energy estimation model for neural network architecture. We propose and evaluate our methodology to estimate energy consumption over different hardwares and experimental set-up. 
%Moreover, we demonstrate that a layer-based model can accurately estimate energy consumption across various tasks, regardless of the specific task and for multiple deep-learning algorithms 
By outperforming state-of-the-art (SOTA) models in terms of accuracy, we demonstrate that a layer-based approach can enhance prediction quality across multiple tasks and generalize to entirely new tasks without requiring re-training, distinguishing it from traditional models.
%(including transformer based algorithms \mla{mla: pkoi le preciser? et si c'est important, j'aurais mis une phrase indépendante pr dire pkoi comme par ex : "tu veux tester plusieurs architectures de LLMs)})  
%\mla{mla: quel type de connaissance? les méthodes SOA ont besoin d'une connaissance a priori?} \adrien{En fait ce que je veux dire c'est qu'on ne considère même pas la tâche dans la modélisation}
% --> Finalement pas dit ...
% Importance de pouvoir s'occuper de plusieurs tâche à la fois car la tâche en elle-même a un impact sur l'énergie dépensée, à cause du type de données géré mais aussi des opérations réalisées Researchers from Hugging Face have benchmarked the electricity usage\cite{luccioni_power_2024}
% [Préciser les apports de l'article par rapport à la décomposition par layers: on est passé à beaucoup plus d'architectures, plus de layers différents mais aussi passé à une étude sur GPU sans contraintes et pas seulement CPU ou GPU fix state > Plus agnostique]
This paper addresses the emerging need for methodologies and tools that enable users to predict the energy consumption of AI services proactively. 
It provides accurate predictions of neural network energy consumption at the layer level, enabling stakeholders to make informed decisions about energy efficiency during AI development and deployment.
% We aim to empower stakeholders to make more sustainable and efficient choices in the design and deployment of AI systems.  
% \mla{mla: mais tu proposes un modèle pour l'EC d'un algo AI mais pas d'un service donc il mq une phrase de transition entre le contexte que tu viens de décrire et la recherche sur la conso energetique des modèles AI}
%
Our key contributions are as follows:

\begin{itemize}
\item \textbf{Experimental protocol} \space %We establish a rigorous experimental protocol to ensure reliable and reproducible measurements when measuring the energy consumption of AI algorithms and the consumption of their individual layers when running on a GPU. 
We establish a rigorous experimental protocol to ensure reliable and reproducible measurements of the energy consumption of AI algorithms and their individual layers on GPUs.
    %This includes analyzing the impact of the number of repetitions on experimental results to ensure reliable and reproducible measurements. Furthermore, we take into account state-of-the-art procedures in energy measurement methodologies to provide a comprehensive and standardized approach. --> Raccourci

\item \textbf{Extensive dataset} \space We collect energy consumption data for neural networks architectures across 3 widely used tasks and 3 hardware with a maximum of 295 architectures, totaling over 100,000 energy measurements, creating one of the most comprehensive datasets for energy estimation in AI considering full-architecture and individual layer consumption.

\item \textbf{Architecture features extraction} We develop a Python-based framework capable of rigorously extracting the layer-wise decomposition of any PyTorch architecture, including the layers and their main characteristics. This enables detailed analysis and modeling of energy consumption at the layer level.

\item \textbf{Layer-wise energy estimation methodology} \space We propose \wattlayer, a task-agnostic methodology to estimate the energy consumption of AI algorithms.By breaking down architectures to a level of granularity sufficiently fine to eliminate task-specific dependencies and unlike SOTA models which require retraining for each task, our approach generalizes across diverse algorithms without customization, achieving superior accuracy and adaptability.
    % \item LLM. \fred{Write something like: Last, we demonstrate that our layer-wise energy model exhibits zero-shot generalization to Large LAnguage Models (LLMs). 
    % Or 
    % Last, we tested the zero-shot generalization capacity of our layer-wise energy model on a small number of Large LAnguage Models (LLMs).  We show that it exhibits good performances (error smaller than 30\% )
    % This suggests that our approach captures fundamental hardware-operation relationships that remain invariant across model scales and architectures paradigms.} 

\item \textbf{Zero-shot generalization to large LLMs} \space Finally, we test the zero-shot generalization capacity of our layer-wise energy model on a small number of LLMs. The model successfully estimates energy consumption for LLMs without fine-tuning (Mean Absolute Percentage Error MAPE $\leq 30\%$), showcasing its generalization capability and the promising potential as a reliable tool for sustainable AI initiatives. %\mla{ajouter ici explication de zero-shot? }

\end{itemize}

\section{Related Work}

Researchers have long worked on energy estimation models, initially for general programs~\cite{dubois_parallel_2012} and more recently for AI algorithms~\cite{rodriguez_evaluating_2024}.
\cite{saborido_impact_2015} approximate the value of the energy consumption by measuring power over discrete intervals $\Delta \tau$ and assuming a constant power between two measurements.  Given a set of $k \in \mathbb{N}$ intervals of length $\Delta \tau \in \mathbb{R}$, $E = \sum_k P(k \cdot \Delta \tau) \Delta \tau$ illustrates the link between the energy consumption $E$ and the power $P$. 
%The associate sampling frequency can be defined as $f_\text{mes} = \frac{1}{\Delta \tau}$.
Starting from this definition, \cite{yang_part-time_2024} seek to understand the internal mechanisms of the energy consumption measurement provided by NVIDIA-smi, a command-line utility provided by NVIDIA that can be used for monitoring, managing, and querying GPU statistics.  
%They assess their results using the most recent NVIDIA's GPU. 
They propose some very useful guidelines to help collect power and energy data through NVIDIA-smi. 
%In particular, they link the energy data collection with the idea of re-assembling a discrete signal.
\cite{li_evaluating_2016} propose an evaluation of the energy consumption of several Convolutional Neural Network (CNN) on GPU and CPU. They evaluate the impact of several experimental setups on the resulting energy consumption including libraries, batch size and hardware settings. They also propose a first evaluation of the energy consumption of the layers of several CNN. They obtain that, among convolutional, pooling, fully connected, and ReLU layers, the convolutional layers are by far the most consuming layers (87\% of the overall energy consumption).

A common way to estimate the energy consumption is to use the number of operations performed as features of the energy consumed by a neural network: FLOPs or MACs~\cite{rodrigues_synergy_2018,goel_energynn_2021,desislavov_trends_2023,getzner_accuracy_2023}. However, this feature alone is often insufficient and requires either supplementing the model with other features or modifying the model itself~\cite{sze_efficient_2017}. 
Whats more, using architecture-level estimation model prevents generalization on different tasks. As our experiments in Section~\ref{sec:results} show, adding new tasks to the database weakens the estimation performance of the final model.
\cite{bytezcom_icml_2025} collect the energy consumption of 1,200 vision models. They propose a log-linear regression %model 
based on 3 hyper-parameters of the model to estimates its consumption: number of parameters, gmacs and activation count. Overall, their model has an error of 34\%. 
%\adrien{This model provides a useful benchmark for evaluating the performance of our estimation model. However, the reported metric—mean absolute percent error—should be interpreted with caution, as it is derived from a regression analysis conducted without separating the database into training and testing datasets. Additionally, the authors do not account for batch size, which our experiments have identified as a significant factor influencing results.} 
While this model offers a robust benchmark and demonstrates strong performance in controlled conditions, its scope is limited to vision architectures. 
%In contrast, our objective is to develop a task-agnostic model. As we demonstrate in Section~\ref{sec:results}, regression models based exclusively on vision networks fail to generalize to other domains. Furthermore, our evaluation protocol strictly separates training and testing sets to assess generalization to unseen architectures, while explicitly incorporating batch size, a critical parameter identified in our experiments (see Appendix~\ref{app:batchsize}).
The authors also propose a second estimation model that utilizes throughput as a feature, delivering exceptional results with an error rate of less than 1\%. They suggest this metric as a viable option for estimation in scenarios where GPU information is unavailable but throughput data is accessible. However, measuring GPU throughput during an experiment is as challenging as measuring energy consumption, which led to the decision not to compare our model with this second approach.
% Another approach consists in estimating the energy consumption through the energy estimation of the neural network's layers. This methods has shown promising results~\cite{cai_neuralpower_2017}\cite{getzner_accuracy_2023}\cite{zhang_ampere_2025} although the experiments were lacking to ensure broader compatibility with various hardware platforms and flexible hardware configuration. 

Another approach estimates the energy consumption of a neural network by modeling the energy of its individual layers. This strategy has shown promising results~\cite{cai_neuralpower_2017,getzner_accuracy_2023,zhang_ampere_2025}, but existing studies remain limited in terms of hardware coverage and execution settings. NeuralPower proposes a polynomial regression framework for CNNs executed on GPUs, but considers only convolutional, fully connected, and pooling layers, and relies on fixed GPU frequency and voltage settings. \cite{getzner_accuracy_2023} introduces a linear regression model based on features such as MACs but focuses on CPU execution and shows reduced accuracy on real-world architectures. These limitations highlight the need for a more general methodology that supports modern GPU-based workloads under realistic hardware configurations.
We summarize SOTA models in Table~\ref{table:sota}.

\adrien{
\begin{table*}[t]
\begin{center}
\caption{SOTA models summary with features, model types and adressed tasks.} 
%\joanna{on peut pas trouver un model type pour MAC et HJ ? Sinon, faut juste mettre layer-wise en haut à la place de model type}}
% \vskip 0.15in
\begin{footnotesize}
\begin{sc}
 \begin{tabular}{|l||c|c|c|c|c||c||c|c|c|}
    \hline
    \multirow{2}{*}{Model} &
      \multicolumn{5}{|c||}{Features} &
      \multicolumn{1}{|c||}{$\!$Layer-$\!$} &
      \multicolumn{3}{|c|}{Task} \\
      \cline{2-6}\cline{8-10}
    & MACs       & $\!$Activations$\!$ & $\!$Parameters$\!$ & $\!$Input$\!$  & $\!$Batch$\!$  & wise & $\!$Vision$\!$ & $\!$NLP$\!$ & $\!$Audio$\!$ \\
    & $\!$or FLOPs$\!$ &             &            & size & size & model &  &  &  \\
    \hline
    NeuralPower \cite{cai_neuralpower_2017}     & $\checkmark$ &   &   & $\checkmark$  & $\checkmark$  & $\checkmark$ &$\checkmark$  &   &  \\
    Getzner \cite{getzner_accuracy_2023}    & $\checkmark$ &   &   &  &  & $\checkmark$ & $\checkmark$  &   &  \\
    %\cite{rodrigues_synergy_2018}   & $\checkmark$ &   &   &   &   &   &$\checkmark$  &   &  \\
    Mac, e.g. \cite{desislavov_trends_2023}    & $\checkmark$ &   &   &   &   &   &$\checkmark$  &   &  \\
    HJ \cite{bytezcom_icml_2025}      & $\checkmark$ & $\checkmark$ & $\checkmark$ &   &    &  &$\checkmark$  &   &  \\
    \hline
    \wattlayer                       & $\checkmark$ & $\checkmark$ &$\checkmark$  &$\checkmark$  &$\checkmark$  &$\checkmark$  &$\checkmark$  &$\checkmark$  & $\checkmark$\\
    \hline
  \end{tabular}
\label{table:sota}
\end{sc}
\end{footnotesize}
\end{center}
% \vskip -0.1in
\end{table*}
}

\section{Methodology and Experimental Set-up}

\subsection{Experimental Protocol}

The experiments conducted concern short time processes. Indeed, the execution of a layer of a neural network, using the PyTorch library, may take less than a second. To ensure high-quality experimental data and improve reproducibility, it is essential to establish a rigorous experimental protocol. This includes defining the data collection tool, the frequency of measurement and the number of repetitions of one execution.

% \begin{itemize}
%     \item The tool used to collect the data ($Tool$)
%     \item The number of repetition of one execution ($N_\text{mes}$)
%     \item The frequency of measurment ($f_\text{mes}$).
% \end{itemize}

\noindent\textbf{Data Collection Tool} \space Several applications are available to measure the energy consumption or power usage of NVIDIA GPUs. %, such as CodeCarbon~\cite{benoit_courty_2024_11171501}, Carbon Tracker~\cite{anthony_carbontracker_2020}, and Scaphandre,\footnote{\href{https://github.com/hubblo-org/scaphandre}{https://github.com/hubblo-org/scaphandre}} among others. 
Most, if not all, of these applications rely on the same tool for estimating NVIDIA GPU energy consumption: NVIDIA-smi. 
\cite{yang_part-time_2024} have investigated the internal mechanisms of energy measurement provided by NVIDIA-smi and found that the tool has an overall measurement error of 5\%. Based on this finding, we adopt NVIDIA-smi for our experiments. Specifically, we use CodeCarbon~\cite{benoit_courty_2024_11171501} to measure energy consumption, which combines Intel RAPL for CPU energy measurement and NVIDIA-smi for GPU energy measurement.

\noindent\textbf{Frequency of Measurement} \space The sampling frequency $f_\text{mes}$ is set to 10 Hz to get closer to the value recommended by \cite{yang_part-time_2024} without raising errors from CodeCarbon.

\noindent\textbf{Repetition} \space 
To obtain reliable energy measurements, the duration of each experiment must be substantially longer than the sampling interval of the power sensor. We therefore repeat each inference multiple times and measure the cumulative energy consumption.
As shown in Section~\ref{subsection:tmes}, the number of repetitions directly affects measurement precision. Based on these experiments, we define a minimum number of forward passes, denoted $N_{mes}$, and set it to 4,000 for most experiments and 15,000 for natural language processing (NLP) layers.

\subsection{Layer-Wise Energy Estimation Framework}

\begin{figure}[tbp, width=0.48\linewidth]
    \centering
    \begin{minipage}[c]{0.48\linewidth}
        %\caption{Illustration of \wattlayer's methodology to estimate the energy of neural networks algorithms.
        % The datasets consist of energy consumption measurements for entire architectures across various configurations, as well as energy data for individual layers. During the training phase, layers are categorized by type, and a dedicated estimation model is developed for each layer type. In the testing phase, architectures are broken down into their constituent layers, and the energy consumption of each layer is predicted using the corresponding layer-specific model. The total energy consumption of the architecture follows by summing the estimated energy values of all layers.}
        \caption{Overview of \wattlayer for neural network energy estimation. Energy measurements are collected for complete architecture and individual layers. During training, layers are grouped by type and a dedicated model is fitted to each group. During inference, a target architecture is decomposed into layers, each layer's energy is estimated with the corresponding model, and the total energy is obtained by aggregating all layer estimates.}
        \label{fig:methodology_illustration}
    \end{minipage}
    \hfill
    \begin{minipage}[c]{0.48\linewidth}
        \centering
        \includegraphics[width=\linewidth]{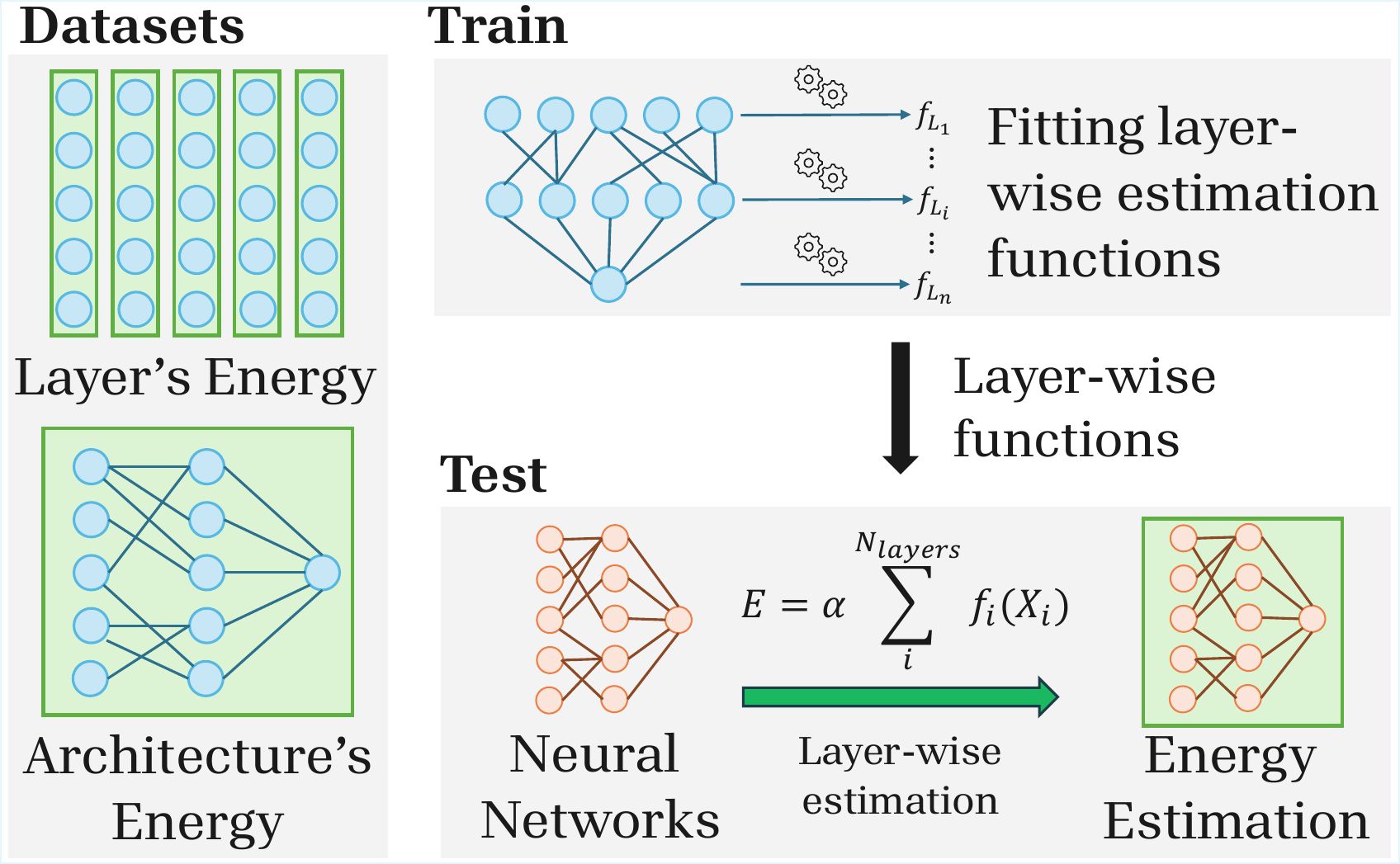}
    \end{minipage}
\end{figure}

\noindent\textbf{Framework} \space 
%Our methodology for estimating the energy consumption of an architecture involves estimating the energy consumption of each individual layer and then summing these values. To achieve this, we train multiple layer-specific estimation functions using the parameters of each layer. These functions allow us to estimate the energy consumption for each layer accurately. Finally, the overall energy consumption of the architecture is determined by aggregating the energy estimates of all layers.
Our methodology estimates the energy consumption of a neural network by decomposing the architecture into individual layers. A dedicated estimation model is trained for each layer type using the parameters of each layer. The total energy consumption of the architecture is then obtained by aggregating the estimated energy of all constituent layers.
Figure~\ref{fig:methodology_illustration} illustrates the methodology schematically.

% %\begin{definition}[
% \textbf{Layer-wise Energy Estimation} \space
% Let $\mathcal{A}$ be a neural network architecture with $N (\in \mathbb{N}$) layers $L_1, ..., L_N$. 
% $E_\mathcal{A}$ and $\lbrace E_{L_i} \rbrace_i$ are respectively the energy consumption of the architecture and of the layers. 
% Let $\lbrace X_i \rbrace_{i \in [|1,N|]}$ be the relevant parameters characterizing the energy consumption of the layers. 

% \fred{We posit that the total energy $E_\mathcal{A}$ can be approximated by the sum of energy functions of its layers, adjusted by a global coefficient $\alpha$}

% \fred{Or: Our model expresses the total energy...} 
% Then $\forall i \in [|1,N|]$, there exists (\fred{I am not sure of the nature of such a statement. It looks a bit like a proposition, however, it may not be really true that the sum is correct. It may be better to write it as our assumption, that we later confirm.})a layer-specific function $f_i$ such that,
%     \begin{align}
%         E_\mathcal{A} \approx \alpha \cdot \sum_{i=1}^N E_{L_i} = \alpha \cdot \sum_{i=1}^N f_i(X_i)~,
%     \end{align}
% \adrien{where $\alpha$ is a correction factor to take into account the gap between architecture energy and layer's energy summation.} 
% \fred{I guess $f_i$ is not really specific to layer $i$, but in fact, we are learning a function for each type of layer with features, e.g. MAC, parameters, etc. Maybe change to $f_{\text{type}_i}(\mathbf{x_i})$ or $f_{t_i}(\mathbf{x_i})$} 
% \fred{Define what are the relevant parameters, maybe add examples. }
% %\end{definition}

\noindent\textbf{Layer-Wise Energy Estimation} \space
Let $\mathcal{A}$ be a neural network architecture with $N (\in \mathbb{N}$) layers $L_1, ..., L_N$. 
$E_\mathcal{A}$ and $\lbrace E_{L_i} \rbrace_i$ are respectively the energy consumption of the architecture and of the layers. We make the assumption that the total energy $E_\mathcal{A}$ may be approximated by the sum of the energy consumption of its layers $\lbrace E_{L_i} \rbrace_i$ up to a correction factor $\alpha$: $E_\mathcal{A} \approx \alpha  \sum_{i=1}^N E_{L_i}$.
% \begin{align}
%    E_\mathcal{A} \approx \alpha  \sum_{i=1}^N E_{L_i}~.
% \end{align}
Despite their large diversity, existing architectures use a limited number of types of layers (e.g. \text{Linear}, \text{Conv2d}, \text{Embedding},...). We note $\text{type}(i)$ the type of layer $i$. %Appendix~\ref{app:layer_types} provides more detail about the layer-types.
Then, for each type, we define a layer-specific function $f_{\text{type}(i)}(\mathbf{x}_i)$ that estimates the energy consumption of layer $i$, where $\mathbf{x}_i$ denotes the relevant parameters impacting the energy consumption of layer $i$. These parameters are: \#MAC, \#Activation, \#Parameters, Input shape, Batch size and Kernel size for convolutional layers (see Appendix~\ref{app:features}). 
%\adrien{More details about the features can be found in Appendix~\ref{app:batchsize}}.
Finally, we write
\begin{align}
    E_\mathcal{A} \approx \alpha  \sum_{i=1}^N E_{L_i} = \alpha  \sum_{i=1}^N f_{\text{type}(i)}(\mathbf{x}_i)~,
    \label{eq:model}
\end{align}
where $\alpha$ is a correction factor to take into account the gap between architecture energy and layer's energy summation.

\noindent\textbf{Data Collection} \space To train the layer estimation function, we need to construct a training dataset of layer energy consumption. For each layer, energy data is collected by running the layer alone using the experimental protocol
with randomly generated tensor. %of the corresponding input size. 
During the evaluation phase, testing data is also generated based on randomly generated input tensor with the same distribution. 
%Data distribution doesn’t differ between training and evaluation.
Associated parameters required for training layer-wise model are also collected.

In their work on estimating the energy consumption of vision neural network layers, \cite{getzner_accuracy_2023} generate random layers to build their training dataset. This approach enabled them to produce sufficient data to train statistical models for various layer types. (e.g., Convolutional, Linear, MaxPooling). However, this method sacrifices real-world layer specificities, resulting in strong performance on random architectures but poorer results on real-world architectures. To address this limitation, we choose to collect data from real-world architectures to build our training dataset. While this approach is more challenging, requiring the extraction of architecture structures without prior knowledge, it provides a more accurate basis for estimating layer energy consumption.
To automate the extraction of neural network structures across multiple architectures, we develop a Python script that sets hooks on each layer. 
A \textit{layer} is defined as a leaf module in the architecture hierarchy. %, i.e., a module that does not contain any child modules. 
The architecture is recursively traversed, following the forward pass of the architecture, and only these terminal modules are retained. This results in a fine-grained decomposition of the network, typically corresponding to elementary operations such as convolutional layers, fully connected layers, activation functions, and normalization layers.
To ensure accurate extraction, it is crucial to account for the actual utilization of layers during inference, rather than relying solely on the layers' names listed in the PyTorch structure (as proposed by \cite{getzner_accuracy_2023}). Our approach captures the true layer decomposition and considers the sequential behavior of layers, including jumps and bypasses that may occur during inference. 
Information regarding our Python scripts can be found in the supplementary materials.
%This script sequentially extracts relevant features during the forward pass. Unlike the method proposed by \cite{getzner_accuracy_2023}, which only extracts layer names, our approach tracks the actual execution phases and accounts for jumps, enabling a detailed layer-wise decomposition of any PyTorch structure given the appropriate input. 
%We provide additional details in Appendix~\ref{app:structure_extraction}.

\noindent\textbf{Model Training} \space The constructed dataset includes energy consumption measurements for architecture layers along with features that characterize energy usage. 
Layers are grouped by type, independently of the architecture or the task from which they were extracted.
This fine-grained, task-independent decomposition enables the aggregation of data from multiple tasks into a larger training set.
%By breaking down architectures to a level of granularity sufficiently fine to eliminate task-specific dependencies, this approach enables the construction of a statistically larger database for training the layer-wise estimation model. 
Ultimately, it aims to enhance prediction quality when transitioning across multiple tasks, rather than compromising it, as is often the case with other models.
A dedicated layer-estimation function is then trained for each group using the collected features. 
To ensure the model remains lightweight and interpretable, we prioritize simple statistical models such as linear regression and multi-linear regression. For specific layers, we also employ log-linear regression similarly to \cite{bytezcom_icml_2025}. 
%Finally, we emphasize that our methodology organizes layers by type, independently of the architecture or the task from which they were extracted. By breaking down architectures to a level of granularity sufficiently fine to eliminate task-specific dependencies, this approach enables the construction of a statistically larger database for training the layer-wise estimation model. Ultimately, it aims to enhance prediction quality when transitioning across multiple tasks, rather than compromising it, as is often the case with other models.

\noindent\textbf{Model Testing} \space To evaluate the performance of our model, we test it on architectures whose energy consumption data was not included in the training dataset. The estimation process for each architecture involves three steps: 
(i) extracting the layer decomposition of the architecture, 
(ii) estimating the energy consumption of each layer using the trained layer-estimation functions, and 
(iii) reconstructing the total energy consumption of the entire architecture by summing the individual layer estimations.

\section{Results}
\label{sec:results}

In this section, we present our results in 4 parts. First, we study the number of executions needed to correctly measure the energy consumption of a process. 
%This number is critical to ensure proper data collection and robust estimation models. 
Next, we assess the layer-wise decomposition hypothesis. We show that a correction factor needs to be applied for our methodology to be accurate.
%and experimentally show that there is a quantitative gap between the energy consumption of an architecture and the summation of its layers' consumption. 
We then evaluate \wattlayer %, our inference energy estimation model, 
on several tasks. We demonstrate that, by leveraging layer-wise energy consumption modeling, it outperforms state-of-the-art models.
Finally, we demonstrate that our methodology paves the way for task-agnostic estimation model by showcasing \wattlayer's ability to accurately estimate the energy consumption of LLMs, despite no model of this family being included in the training dataset. 
%Information regarding our Python scripts is available in Appendix~\ref{app:structure_extraction} and supplementary material.

\subsection{Impact of Measurement Time}% on Experimental Results}

\label{subsection:tmes}

\begin{figure}[htbp]
    \centering
    \includegraphics[width=0.4\linewidth]{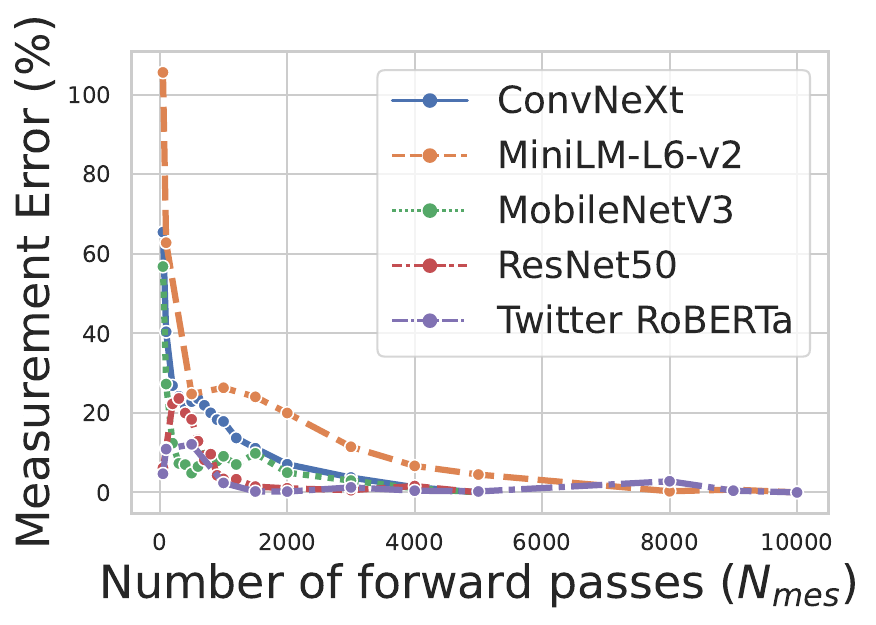}\hspace{0.05\linewidth}%
    \includegraphics[width=0.4\linewidth]{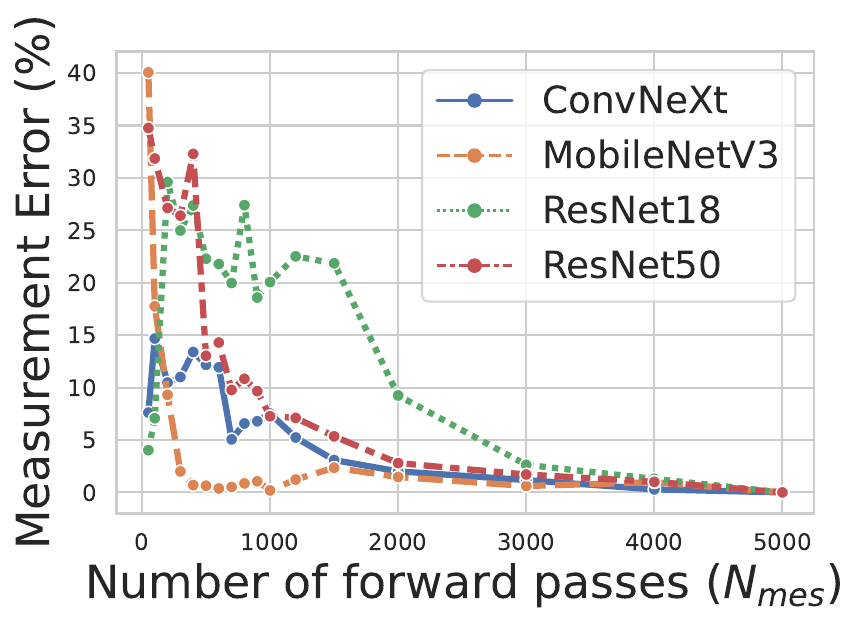}
    \caption{$\text{Error}(N_\text{mes})$ with respect to the number of repetitions of a forward pass $N_\text{mes}$. Experiments ran on an NVIDIA GPU RTX 6000 (left) and NVIDIA GPU TITAN X (right) with batch size equal to 1.}
    \label{fig:N_mes_err}

    \vspace{8pt}

    \includegraphics[width=0.4\linewidth]{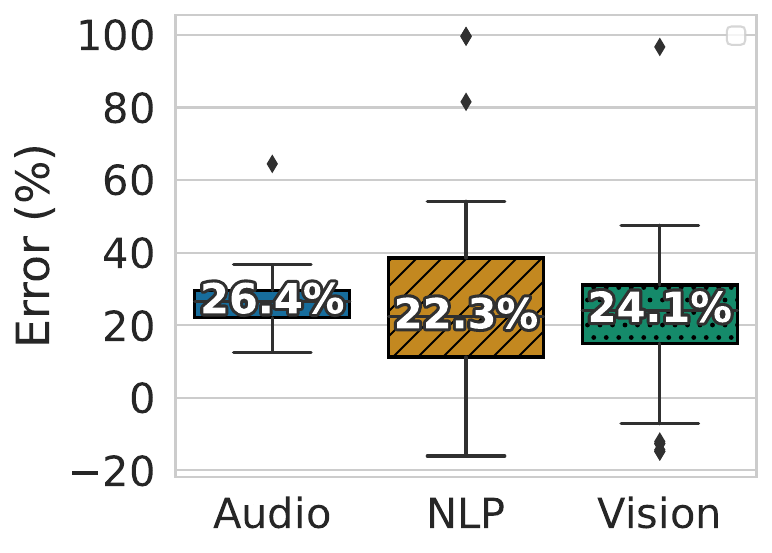}\hspace{0.05\linewidth}%
    \includegraphics[width=0.4\linewidth]{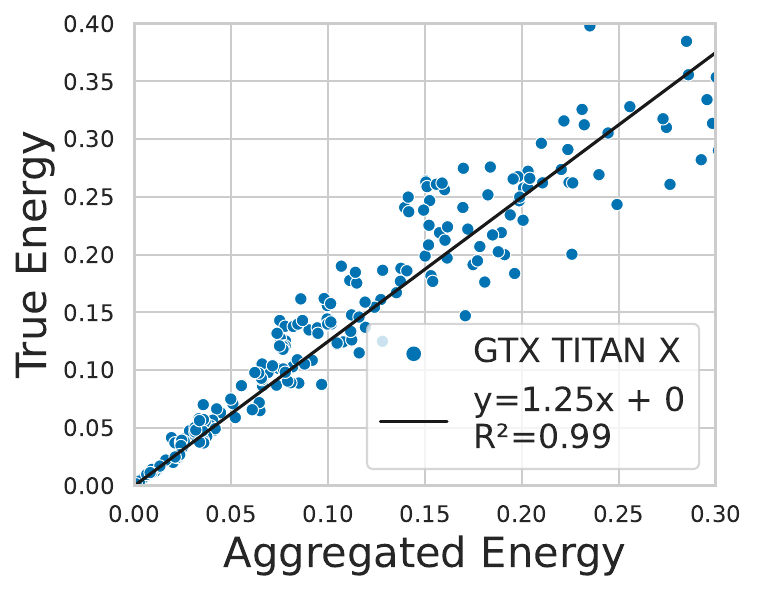}
    \caption{Layer decomposition error (left) and correction factor calibration (right).}
    \label{fig:boxplot_layer_assumption}
\end{figure}

% To ensure that the experiment runs long enough for the measurement tool to provide an accurate approximation of the energy consumption, %for a single forward pass.
% To obtain a reliable approximation of the energy consumed during one execution, 
We repeat the forward passes $N_\text{mes}$ times and compute the average energy consumption: %by dividing the total measured energy by $N_\text{mes}$: 
$E_\text{collected} = {E_\text{experiment}}/{N_\text{mes}}$.
By defining the experimental error as: 
$\text{Error}(N_\text{mes}) = \frac{|E_\text{collected}(N_\text{mes}) - E_\text{collected}(N_{\max})|}{E_\text{collected}(N_{\max})} $
we set $N_\text{mes}$ to the minimum value that guarantees an error lower than a desired threshold.
We evaluate $N_\text{mes}$ %using the framework described in Algorithm~\ref{alg:n_mes} 
for 2 tasks and on 2 GPUs (see Figure~\ref{fig:N_mes_err}).
We demonstrate that a minimum value of $N_\text{mes}$ is required to ensure an error below 20\%. 
%(see Appendix~\ref{app:t_mes}). 
Furthermore, setting $N_\text{mes}$ correctly depends on the process that is run. If a specific process requires only a few calculation steps, then $N_\text{mes}$ needs to be set to a larger value.

%% Combined figure: Figure 2 (Error(N_mes)) stacked on top of Figure 3 (Layer
%% Decomposition Error / Correction factor). Within each figure the two plots are
%% placed side-by-side at 0.4\linewidth and centered as a pair (a fixed gap plus
%% \centering gives equal left/right margins, instead of \hfill spreading them to
%% the edges). Each figure carries its own \caption so the two separate figure
%% numbers and labels are preserved. The previous Figure 3 block (further below)
%% has been merged here. The ceurart caption \parbox uses the default \linewidth.

\subsection{Assessing the Layer-Wise Decomposition Hypothesis}

\label{Layer-wise_decomposition_hypothesis}

Our methodology relies on the assumption that the energy of an architecture can be approximated by the aggregation of the energy consumption of its layers (see Equation~\ref{eq:model}). Our experimental findings, detailed in Figure~\ref{fig:boxplot_layer_assumption}, show that a simple summation of the energy consumption of layers tends to underestimate the total consumption by approximately 25\%. Different from~\cite{getzner_accuracy_2023} findings, we attribute this discrepancy 
to measurement granularity and system-level overheads, such as memory management, data movement or GPU frequency setting, that are not captured when profiling layers in isolation. 
Consequently, we calibrate the \wattlayer correction factor $\alpha$ in \eqref{eq:model} to $1.25$. 
\adrien{
In our experiments, this adjustment is stable across hardware platform and tasks. Moreover, it}
yields robust predictive performance while preserving the model's universality and adaptability to new tasks, 
\adrien{
as demonstrated throughout the subsequent results.}
% this result is most probably due to  , e.g. memory management, not taken into account when running layers independently.
% As a result, the \wattlayer's correction factor of Equation~\ref{eq:model}, $\alpha$, is set to a value of $1.25$, leading to good experimental results while offering universality and ease of adaptation to other tasks, as demonstrated below.  
%The use of a layer-based model remains relevant, as it provides good experimental results while offering universality and ease of adaptation to other tasks, as demonstrated below.  Section~\ref{sec:results}.

%% NOTE: this figure (fig:boxplot_layer_assumption) has been merged into the
%% combined figure in Section "Impact of Measurement Time" above, where it is
%% displayed in the right-hand column next to Figure~\ref{fig:N_mes_err}.

\subsection{Statistical Model to Estimate Layer-Wise Energy Consumption}

\textbf{Evaluation Framework} \space The training dataset was built by collecting the energy consumption data for 92 image-classification models from Timm~\cite{rw2019timm} and 29 text-classification models and 29 audio-classification models from Transformers~\cite{wolf-etal-2020-transformers} across three batch sizes: 1, 32, and 64 for a total of 164 models and 99,411 samples. 
Most parameters were kept at their default values, corresponding to the standard use case. In Hugging Face workflows, models are typically initialized from pretrained configurations, and users generally rely on these default settings unless they explicitly specify alternative values.
To validate our model, we test it on 93 image-classification models from Timm and TorchVision~\cite{torchvision2016}, 17 text-classification and 35 audio-classification models from Transformers that were excluded from the training dataset for a total of 145 models and 207 samples. 
To evaluate accuracy, we use four metrics: mean absolute percent error (MAPE), median absolute percent error (MedAPE), maximum absolute percent error (MaxAPE), and minimum absolute percent error (MinAPE). 

\begin{figure*}[htbp]
    \centering
    \begin{subfigure}[b]{0.33\textwidth}
        \centering\includegraphics[width=1\linewidth]{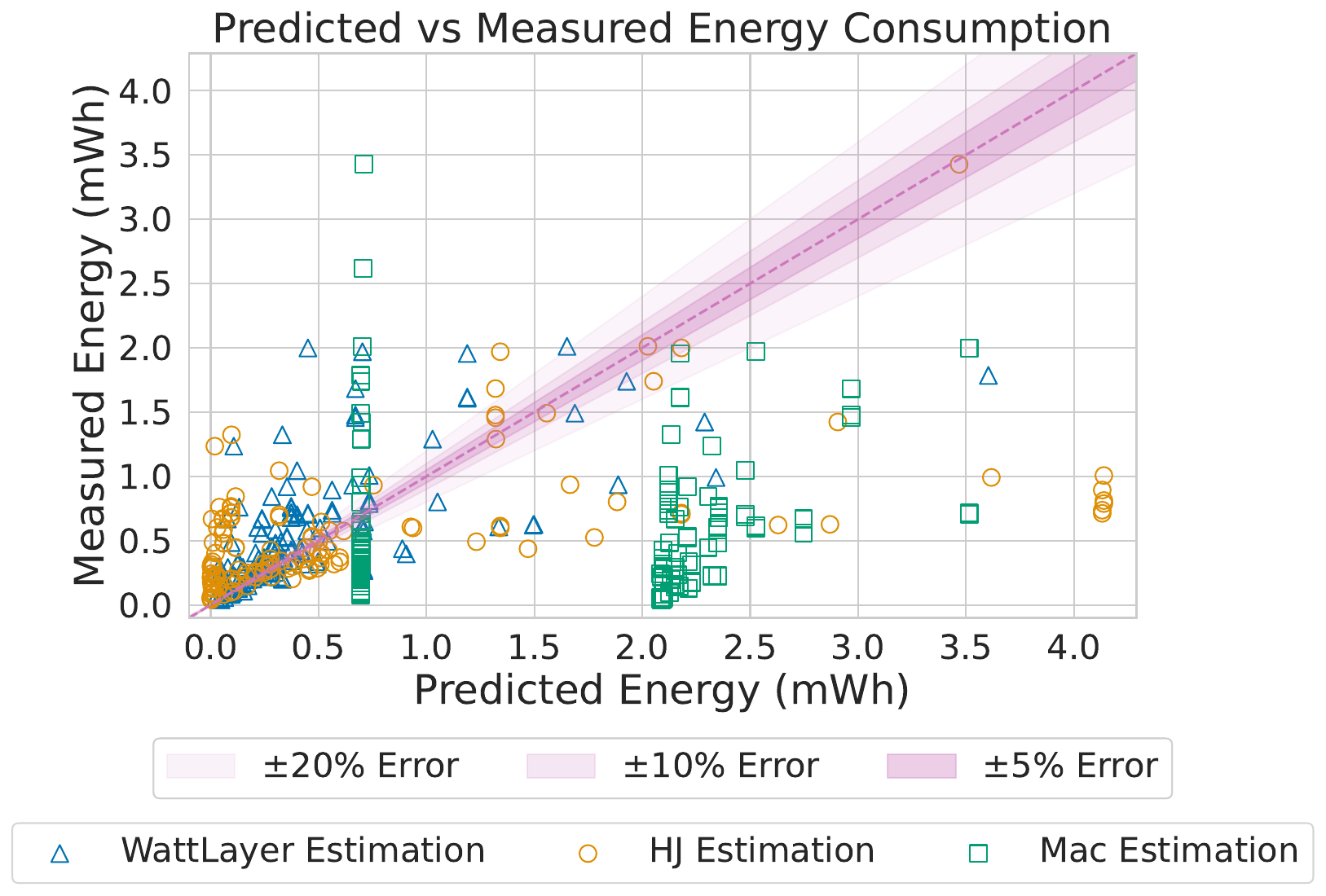}
        \caption{%Predicted vs. measured energy for 
        94 vision architectures}
    \end{subfigure}
    \hfill
    \begin{subfigure}[b]{0.33\textwidth}
        \centering\includegraphics[width=1\linewidth]{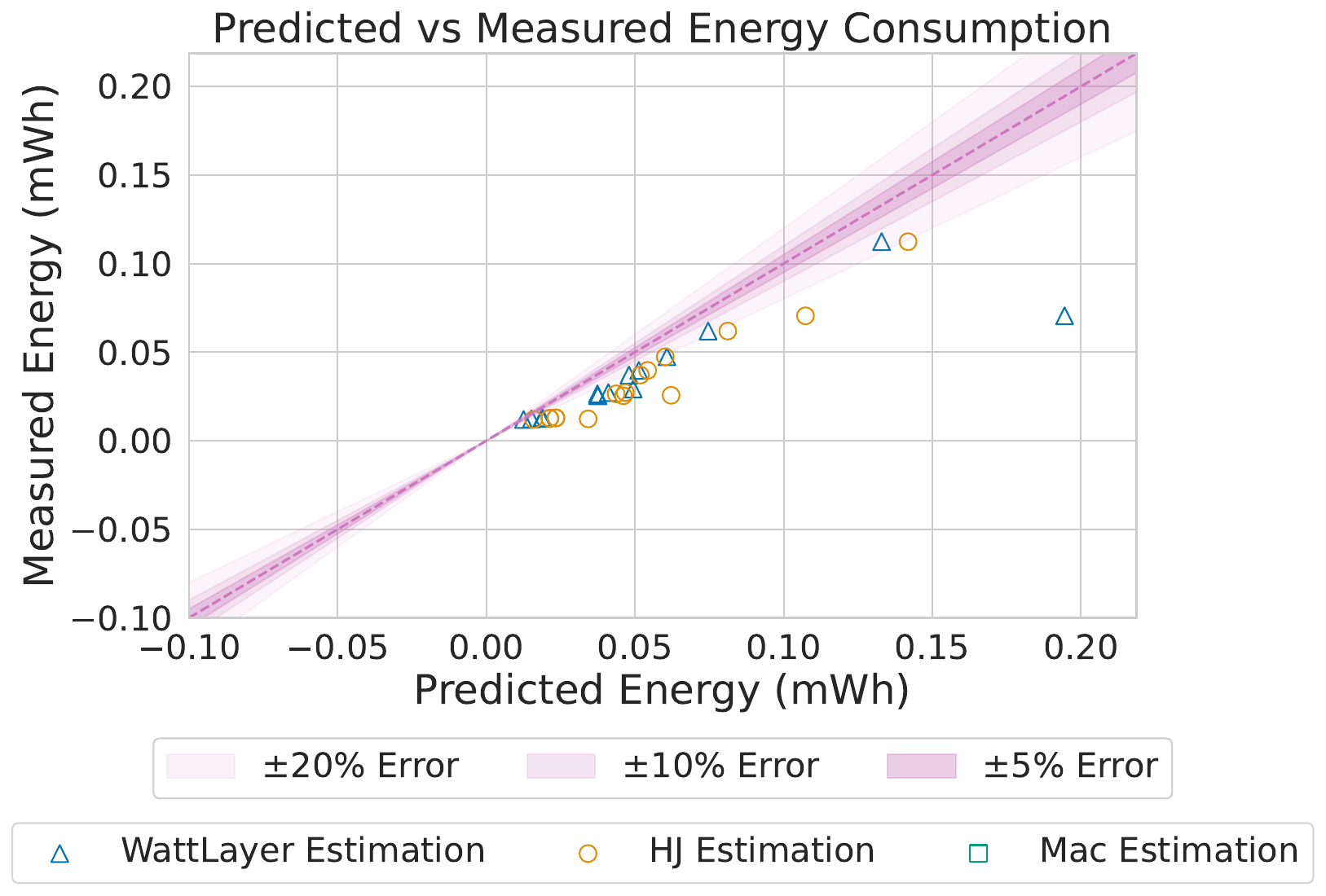}
        \caption{%Predicted vs. measured energy for 
        17 NLP architectures}
    \end{subfigure}
    \hfill
    \begin{subfigure}[b]{0.33\textwidth}
        \centering\includegraphics[width=1\linewidth]{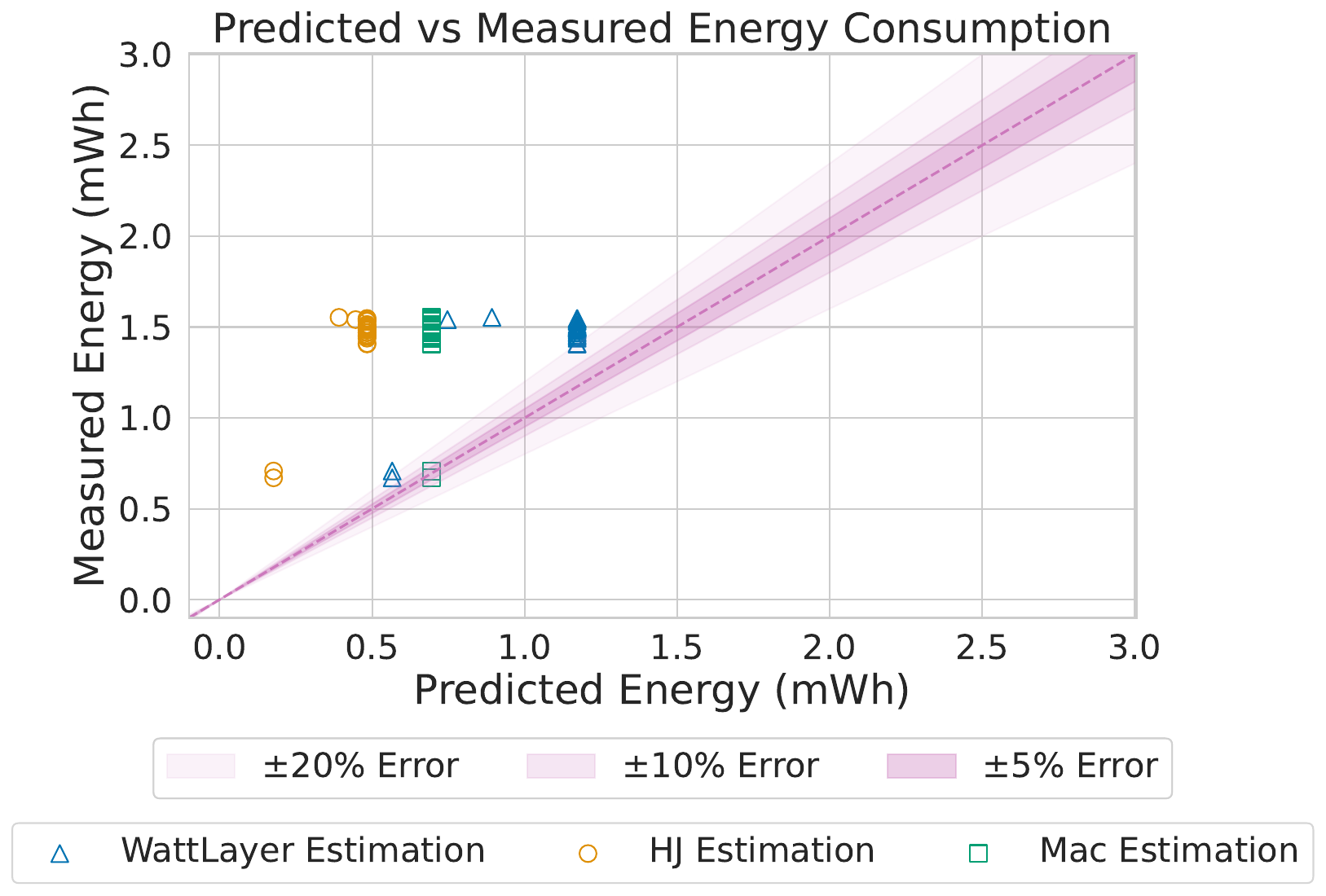}
        \caption{%Predicted vs. measured energy for 
        35 audio architectures}        
    \end{subfigure}
    \caption{Measured vs. predicted energy with \wattlayer, HJ, and Mac estimation models for (a) Vision, (b) NLP and (c) Audio architectures. The evaluation is conducted on architectures sourced from the widely used Python libraries Torchvision and Transformers.}
    \label{fig:prediction_vs_measured}
\end{figure*}

\noindent\textbf{Baselines} \space 
We compare \wattlayer against two strong and widely applicable baselines. The first is a linear regression model using the number of MACs as the sole predictor, a common approach in prior work and practice. The second is the state-of-the-art ``HJ'' model proposed by \cite{bytezcom_icml_2025}, which employs a log-linear regression based on three architectural features: \#MACs, \#Parameters, and \#Activations.
 We do not directly compare against earlier layer-wise methods as their assumptions limit applicability in our setting. 
 NeuralPower is restricted to a fixed GPU configuration and a three layer types, limiting its applicability beyond vision models.
 \cite{getzner_accuracy_2023} supports a wider range of layer types but relies on randomly generated layer configurations and extracts layers from the static PyTorch module hierarchy rather than the effective execution graph, limiting its ability to handle modern architectures with skip connections such as ResNet~\shortcite{he_resnet_2015}. 
 In contrast, \wattlayer is based on effective forward-pass execution and a unified layer-wise model covering diverse layer types and tasks, enabling scalable and accurate estimation across architectures and hardware (see Appendix~\ref{app:sota}).
To have a fair comparison we create a balanced train dataset with 25 architectures from each task with which we train both SOTA models (MAC-balanced, HJ-balanced) for all tasks. In addition, for vision only, we also train specific models called HJ-Vision and MAC-Vision because the number of architectures was sufficient to have a fair comparison between our models and the SOTA models. Having those different models ensures an evaluation of good quality and it emphasises the adaptability of \wattlayer.
% Training metrics are summarized in Table~\ref{table:hj_mac_stat} and 
Figure~\ref{fig:prediction_vs_measured} provides a visualization of the model performance against the two SOTA models. 
\begin{table*}[tbp]
\caption{Performance metrics of \wattlayer compared to SOTA models on image-, text- and audio-classification models.}
\label{table:results-image-class}
\label{table:results-text&audio-class}
% \vskip 0.15in
\begin{center}
\begin{footnotesize}
\begin{sc}
\begin{tabular*}{\textwidth}{@{\extracolsep{\fill}}lrrrrr}
\toprule
Metric & \wattlayer &  \multicolumn{1}{c}{HJ-V} & \multicolumn{1}{c}{HJ-B} & MAC-V & MAC-B\\
\midrule
\multicolumn{6}{c}{Testing on 93 image-classification models}\\
\midrule
MAPE&$\mathbf{40.1\%}$&$63.9\%$&$99.8\%$&$132.6\%$ &$833.7\%$\\
MedAPE&$\mathbf{33.6\%}$&$48.7\%$&$86.8\%$&$52.9\%$&$229.0\%$\\
MaxAPE&$\mathbf{168.9\%}$&$\!\!\!\!259.2\%$&$\!\!\!\!870.1\%$&$\!\!\!\!\!\!3,073.3\%$&$\!\!\!\!18784.3\%$\\
MinAPE&$0.4\%$&$1.9\%$&$\mathbf{0.02\%}$&$0.7\%$&$8.2\%$\\
Std dev&$\mathbf{35.1\%}$&$56.7\%$&$130.4\%$&$353.7\%$ &$2199.7\%$\\
\midrule
\multicolumn{6}{c}{Testing on 17 text-classification models}\\
\midrule
MAPE&$\mathbf{45.6\%}$&$341.4\%$&$152.9\%$&--&$3,091.2\%$\\
MedAPE&$\mathbf{46.2\%}$&$265.4\%$&$68.1\%$&--&$2,526.3\%$\\
MaxAPE&$\mathbf{175.8\%}$&$1,158.1\%$&$1,510.5\%$&--&$5,713.4\%$\\
MinAPE&$\mathbf{4.7\%}$&$105.5\%$&$26.2\%$&--&$518.6\%$\\
Std dev&$\mathbf{35.9\%}$&$320.9\%$&$341.7\%$&--&$1832.8\%$\\
\midrule
\multicolumn{6}{c}{Testing on 43 audio-classification models}\\
\midrule
MAPE&$\mathbf{26.6\%}$&$74.6\%$&$67.7\%$&--&$58.3\%$\\
MedAPE&$\mathbf{22.2\%}$&$68.6\%$&$67.5\%$&--&$53.8\%$\\
MaxAPE&$\mathbf{61.1\%}$&$100$\%&$75.1\%$&--&$93.6\%$\\
MinAPE&$15.6\%$&$64.1\%$&$63.4\%$&--&$\mathbf{1.72\%}$\\
Std dev&$10.1\%$&$12.6\%$&$\mathbf{3.0}\%$&--&$23.7\%$\\
\bottomrule
\end{tabular*}
\end{sc}
\end{footnotesize}
\end{center}
% \vskip -0.1in
\end{table*}

\begin{figure}[tbp]
    \begin{center}
    \includegraphics[width=0.7\linewidth]{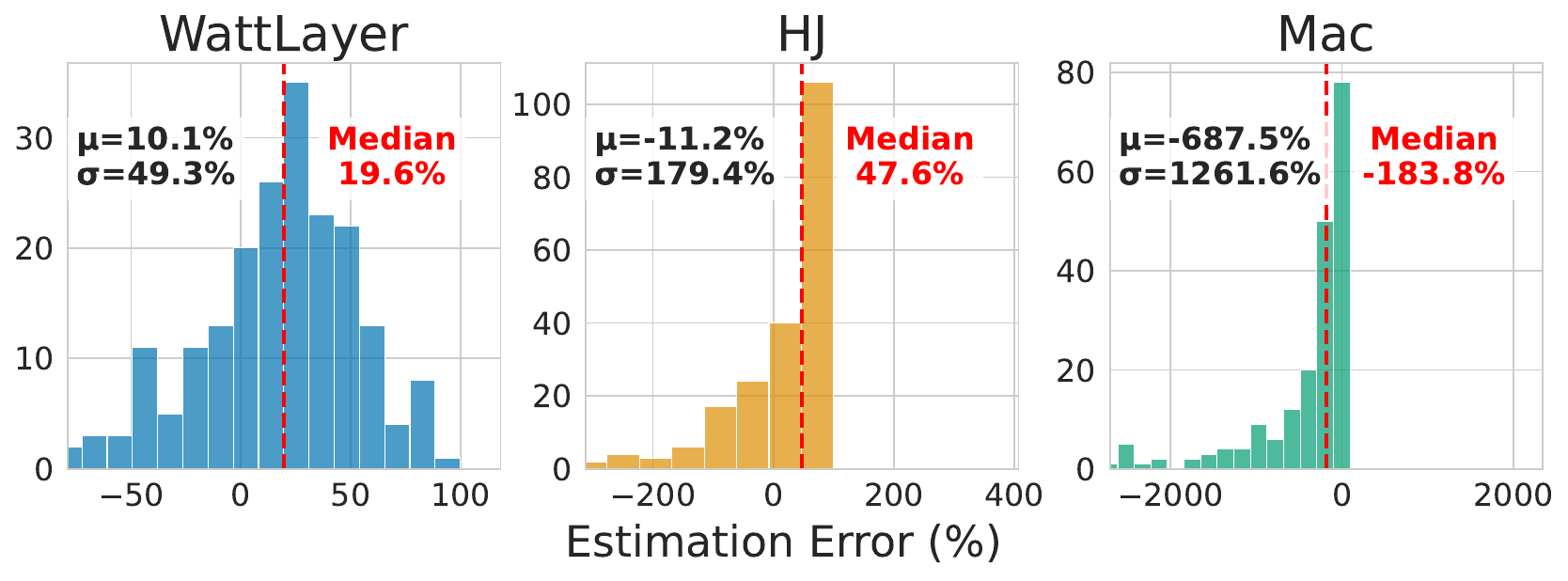}
    \end{center}
    \caption{Distribution of estimation error for \wattlayer, HJ and Mac model across all architectures in the test dataset. The evaluation is conducted on architectures sourced from the widely used Python libraries: TorchVision, Timm and Transformers.}
    \label{fig:distribution_error}
\end{figure}

\noindent\textbf{Vision Models} \space 
We train statistical models for each layer type to predict energy consumption. Three types of models are evaluated: linear regression, multi-linear regression, and log-linear regression. Linear regression is applied to most layer types due to its simplicity and effectiveness. However, for layers that are highly represented in the dataset or accounted for the majority of energy consumption, more complex models, such as multi-linear and log-linear regression, are employed to enhance prediction accuracy. 
%The results of these layer-wise models are summarized in Table~\ref{table:image-class_stat}, demonstrating their effectiveness in modeling energy consumption.
% \begin{figure}
%     \centering
%     \includegraphics[width=0.9\linewidth]{Images/Energy_per_Image_vs_Batch_Size.pdf}
%     \caption{Energy consumption of 3 Transformers architectures for several batch sizes.}
%     \label{fig:batch_size}
% \end{figure}
Results, presented in Table~\ref{table:results-image-class}, show that \textbf{our model outperforms both models with MAPE = 40.1\% and MedAPE = 33.6\%} due to several factors: 
(i) our approach incorporates more specific parameters, enabling finer-grained layer-wise decomposition for improved accuracy; 
(ii) batch size, a critical factor influencing energy consumption per image, is explicitly considered, whereas other models overlook this dependency despite prior findings by \cite{bytezcom_icml_2025} that larger batch sizes reduce energy consumption per image. %(see Figure~\ref{fig:batch_size});
and (iii) while training datasets use 100 different models, state-of-the-art models use %yields 
only 300 samples by focusing on the energy consumption of the full architecture (100 models $\times$ 3 batch sizes), whereas our framework generates thousands of samples by also accounting for the energy consumption of individual layers.

\noindent\textbf{NLP and Audio} \space We extend our energy estimation framework to text and audio classification models by jointly training on Vision, NLP, and Audio datasets. This multi-domain setup enables learning of shared layer-wise energy patterns across heterogeneous tasks, moving towards a task-agnostic formulation of energy estimation. Unlike prior work, which typically requires task-specific training or retraining (Table~\ref{table:results-text&audio-class}), our approach relies on a unified layer-wise model applicable across domains.
By exploiting common layer types across modalities (e.g., linear layers in vision, NLP, and audio, or convolutional layers in vision and audio), the model learns transferable energy functions without task-specific redesign. This generalization capability not only simplifies the estimation process but also ensures superior performance and scalability across various applications.
We observe that \textbf{our methodology outperforms SOTA models with MAPE = 46.6\% and MedAPE 46.2\% for text-classification and with MAPE = 26.6\% and MedAPE = 22.2\% for audio-classification}.
Figure~\ref{fig:distribution_error} provides a detailed visualization of the error distribution across the three estimation models. 
Notably, for all tasks, \wattlayer exhibits a significantly more centralized error distribution, with values closer to zero compared to the other models.

\begin{figure}[htbp, width=0.49\linewidth]
    \centering
    \begin{minipage}[t]{0.49\linewidth}
        \centering
        \includegraphics[width=\linewidth]{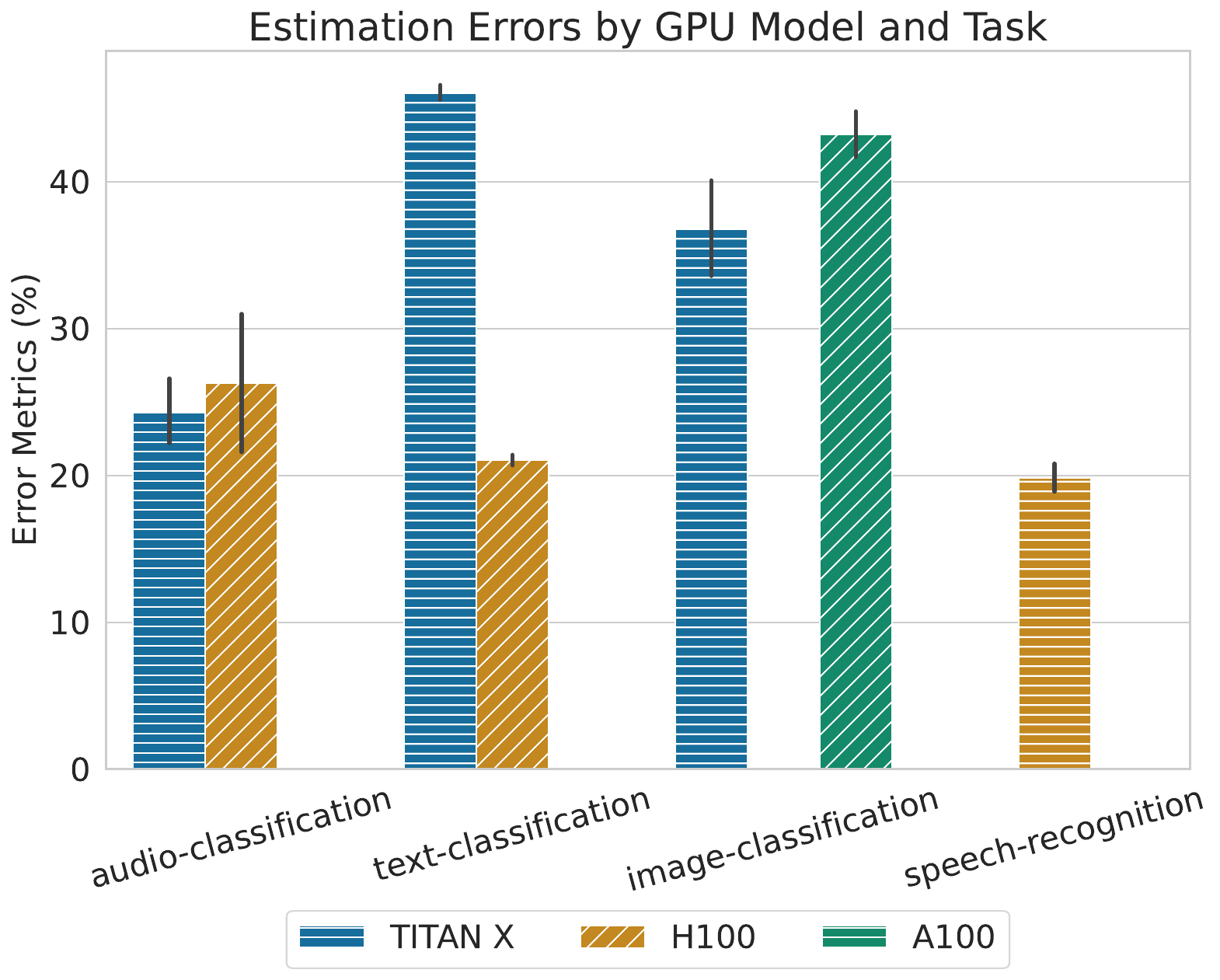}
        \caption{\wattlayer performance for other GPUs: NVIDIA H100 (5 NLP architectures and 18 audio architectures are used for training and 116 models across NLP and Audio for testing) and A100 (19 Vision architectures are used for training and 19 for testing).}
        \label{fig:other_gpu}
    \end{minipage}
    \hfill
    \begin{minipage}[t]{0.49\linewidth}
        \centering
        \includegraphics[width=.9\linewidth]{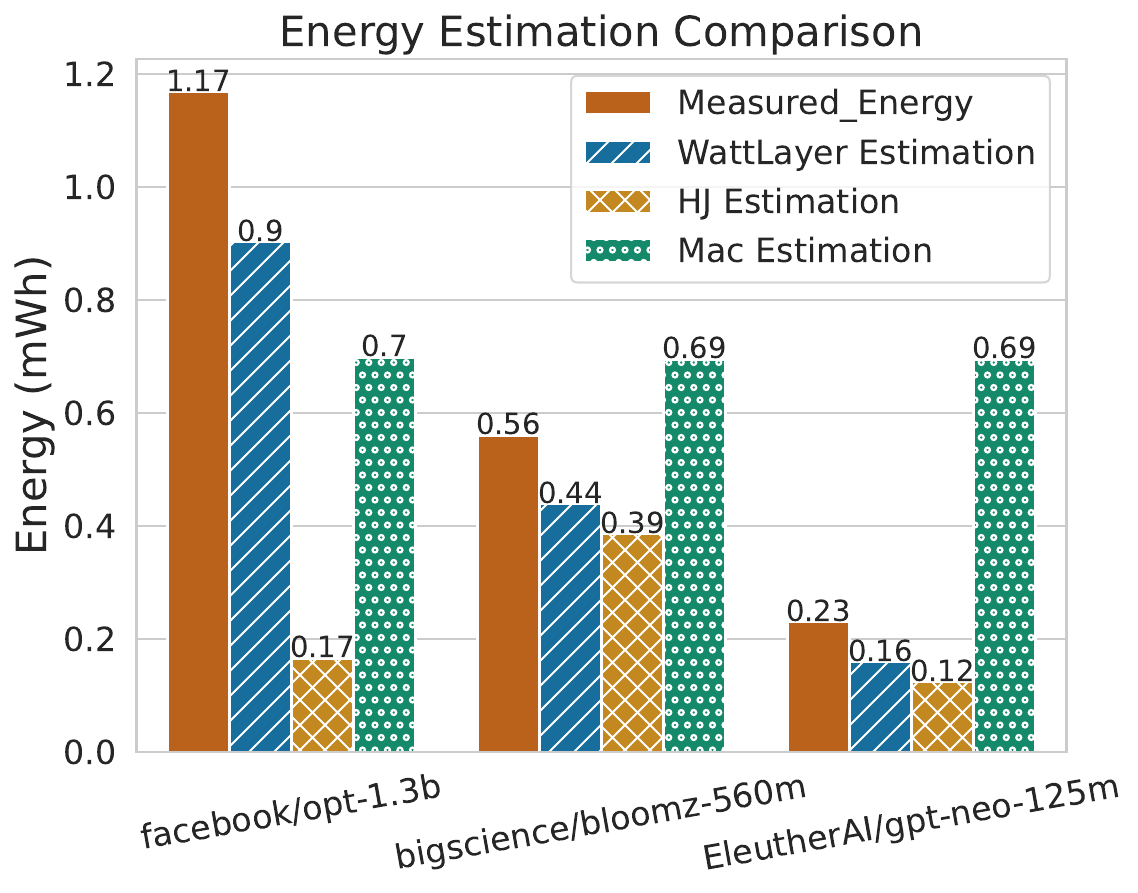}
        \caption{Energy estimation for 3 LLMs architectures facebook/opt-1.3b (1.3B parameters), bigscience/bloomz-560m (560M parameters), and EleutherAI/gpt-neo-125m (125M parameters) using \wattlayer and SOTA models on GPU TITAN X for batch size equal to 1.}
        \label{fig:LLMs}
    \end{minipage}
\end{figure}

% \newpage
Finally, we test our methodology on other GPUs from NVIDIA: H100 and A100 to assess the adaptability of our model on other hardware. We show that \wattlayer has similar MAPE on other hardware platforms (see Figure~\ref{fig:other_gpu}).

%% Combined figure: Figure 6 (other-GPU performance) in the left column and
%% Figure 7 (LLM energy estimation) in the right column, each in its own minipage
%% with its own caption. width=0.49\linewidth keeps both captions within their
%% columns (the ceurart caption parbox uses the figure "width" key, not the
%% minipage). The previous Figure 7 block (further below) has been merged here.

\subsection{Evaluating Zero-Shot Generalization of \wattlayer to LLMs} 

%\fred{question: are the LLMs a lot larger than the models we tested before for the other tasks?}

\label{LLM}

After validating that \wattlayer accurately predicts the energy consumption of unseen architectures within its training domain, we evaluate its zero-shot generalization to large language models (LLMs), without any fine-tuning or LLM-specific training data. This experiment is particularly relevant because LLMs are widely deployed in real-world applications and are known for their substantial energy demands. Collecting dedicated energy measurements for LLMs is costly, as it requires repeated inference of computationally intensive models.

In this setting, the model relies exclusively on the layer-wise estimation functions learned during joint training on vision, text-classification, and audio-classification architectures. No data from text-generation models is included in the training set. We apply the framework to three Transformer-based models from the Transformers library: facebook/opt-1.3b~\cite{zhang_opt_2022} (1.3 billion parameters), bigscience/bloomz-560m~\cite{muennighoff2022crosslingual} (560 million parameters), and EleutherAI/gpt-neo-125m~\cite{gpt-neo,gao2020pile} (125 million parameters). Each model generates text from a single prompt, and measurements are repeated according to the protocol described in Section~\ref{subsection:tmes}.

The results, presented in Figure~\ref{fig:LLMs}, demonstrate that the layer-wise formulation transfers effectively to this previously unseen family of architectures. The estimates leverage layer types encountered during training, including linear layers shared across Vision, NLP, and Audio, embedding layers from text-classification models, and one-dimensional convolutions commonly used in Audio architectures. 
These findings suggest that the proposed framework captures fundamental hardware-operation relationships that remain invariant across model scales and architectures paradigms.
Furthermore, \wattlayer outperforms SOTA models, showcasing its potential as a reliable tool for sustainable AI initiatives.

%% NOTE: this figure (fig:LLMs) has been merged into the combined figure in the
%% previous subsection, where it is displayed in the right-hand column next to
%% Figure~\ref{fig:other_gpu}.

\section{Discussion}

The proposed estimation model offers several notable advantages. Its modular and flexible design allows for the seamless addition of new parameters, enabling continuous improvement of the method.
% It provides accurate predictions of neural network energy consumption at the layer level, enabling stakeholders to make informed decisions about energy efficiency during AI development and deployment.
Furthermore, if new types of layers that significantly impact energy consumption are identified in the future, then specific data can be collected, and a new model be trained to account for these layers. %This adaptability ensures that the method remains relevant and capable of evolving alongside advancements in neural network architectures.
This adaptability ensures that the method remains relevant and can evolve alongside advances in neural network architectures.
Additionally, the lightweight nature of the model ensures that its own energy consumption remains minimal, making it efficient and suitable for practical use.
%
% While the methodology has been tested on several hardware configurations, the model is currently designed to operate on only one machine at a time. 
% While the proposed methodology already demonstrates strong performance across several hardware configurations, it is currently calibrated for one target machine at a time. This design enables accurate predictions by tailoring the model to the characteristics of a specific GPU. Extending this approach to additional platforms only requires collecting calibration data, as demonstrated in our cross-GPU experiments.
% This restricts scalability and applicability in distributed systems. Moving forward, we aim to adapt and upgrade the model to support distributed systems, enabling energy estimation across multiple machines simultaneously. 
%

%Despite these strengths, certain limitations must be acknowledged.
To contextualize these advantages, certain boundaries of the current scope should be noted.
While the proposed methodology already demonstrates strong performance across several hardware configurations, it is currently trained for one target machine at a time. This design enables accurate predictions by adapting the model to the characteristics of a specific GPU. Extending the approach to a new platform requires collecting a relatively small amount of additional data, as demonstrated in our cross-GPU experiments. 
Future work will focus on improving hardware generalization by incorporating GPU-specific descriptors—such as architectural and performance characteristics—similarly to the approach proposed by \cite{mayaki_modeling_2025}. 
Such features could reduce the amount of data required when targeting new devices,
\adrien{improve the estimation of the correction factor $\alpha$ across hardware platforms,}
and move the framework toward a hardware-agnostic formulation. 
\adrien{A broader cross-hardware evaluation of $\alpha$ will also be conducted as part of this effort.}
In addition, we plan to extend the methodology to distributed systems, enabling energy estimation across multiple machines simultaneously.

Although pruning is accounted for in the model due to its impact on layer shapes, other acceleration techniques such as quantization or sparsification have not been explicitly included in the methodology. 
Our primary objective is to derive estimation models that are applicable to the widest range of real-world use cases, in particular those relying on widely adopted libraries such as Transformers and Timm, executed on commodity GPUs with workflows that often includes by default the available accelerations.
Considering specialized hardware that can skip zero-valued computations or incorporating quantization as a feature in future iterations of the model could further enhance its accuracy and applicability but falls outside the scope of our work.
Looking ahead, we also plan to extend our results for more LLMs architectures but also on other generative algorithms, as those models are known to be very energy consuming.

% \mla{ajouter une ou deux perspectives par exemple étendre l'étude de la généralisation sur plusieurs types d'architectures de modèles génératifs autres que les LLMs ... car Y'a pas que les LLM's dans la vie :-)}

\section{Conclusion}

We proposed a task independent layer-wise methodology to estimate the energy consumption of AI algorithms. Unlike state-of-the-art models which require specific training or fine-tuning for each task, our approach generalizes across diverse algorithms without additional customization. By incorporating finer-grained layer-wise decomposition, considering critical factors such as batch size, and leveraging a larger dataset for training, our methodology achieves superior accuracy and adaptability across various AI tasks and architectures.

By addressing the lack of standardized tools for energy estimation, \wattlayer contributes to the advancement of sustainable AI practices. It empowers developers and organizations to proactively evaluate the energy impact of their models, fostering a culture of transparency and accountability in AI design. 
% Furthermore, the methodology presented in this work demonstrates compatibility across diverse hardware configurations and deep-learning architectures, ensuring broad applicability and scalability.
%
This research has the potential to influence both academic and industrial approaches to AI energy management, encouraging the integration of energy efficiency considerations into the life cycle of AI systems. By bridging the gap between energy estimation and practical implementation, \wattlayer supports the global effort to reduce the environmental footprint of AI technologies while maintaining innovation and performance.
%\newline

% \section{Impact Statement}

% The rapid adoption of Artificial Intelligence (AI) across industries has brought transformative benefits but also significant environmental concerns due to the growing energy demands of AI systems. This paper introduces \wattlayer, a task-independent energy estimation model that provides accurate predictions of neural network energy consumption at the layer level, enabling stakeholders to make informed decisions about energy efficiency during AI development and deployment.

% By addressing the lack of standardized tools for energy estimation, \wattlayer contributes to the advancement of sustainable AI practices. It empowers developers and organizations to proactively evaluate the energy impact of their models, fostering a culture of transparency and accountability in AI design. Furthermore, the methodology presented in this work demonstrates compatibility across diverse hardware configurations and deep-learning architectures, ensuring broad applicability and scalability.

% This research has the potential to influence both academic and industrial approaches to AI energy management, encouraging the integration of energy efficiency considerations into the life cycle of AI systems. By bridging the gap between energy estimation and practical implementation, \wattlayer supports the global effort to reduce the environmental footprint of AI technologies while maintaining innovation and performance.

\begin{acknowledgments}

This work is part of the SmartNet challenge project within the Inria–Nokia Joint Lab and has been supported by the French government National Research Agency (ANR) through the UCA JEDI (ANR-15-IDEX-01) and EUR DS4H (ANR-17-EURE-004), by the France 2030 program under Grant Agreements Nos. ANR-23-PECL-0003 and ANR-22-PEFT-0002, through the 3IA Côte d’Azur Investments in the Future project with reference number ANR-23-IACL-0001, and by the European Network of Excellence dAIEDGE under Grant Agreement No. 101120726.
    
\end{acknowledgments}

%% -------------------------------------------------------------------------
%% CEUR-WS requires a "Declaration on Generative AI" (https://ceur-ws.org/GenAI/).
%% This was not present in the original manuscript and only the author can state
%% the actual AI usage, so it is left as a commented placeholder to be filled in
%% (uncomment and complete the relevant option):
%
\section*{Declaration on Generative AI}
% The author(s) have not employed any Generative AI tools.
%
% % Or, using the activity taxonomy in https://ceur-ws.org/genai-tax.html:
During the preparation of this work, the authors used GPT5.1 in order to: Grammar and spelling check.
% % <activity>. 
After using this tool, the authors reviewed and edited the content as needed and take full responsibility for the publication's content.
%% -------------------------------------------------------------------------

%% The bibliography style is set by the ceurart class (elsarticle-num-names);
%% the original \bibliographystyle{named} line has been removed accordingly.
\bibliography{sure26}

\appendix

\section{Features of the Estimation Model}

\label{app:features}

\noindent\textbf{\#MACs (Multiply--Accumulate Operations).}
The number of MACs measures the computational complexity of a layer and is computed using \texttt{ptflops}~\cite{ptflops}. It captures the number of arithmetic operations performed during inference.
\noindent\textbf{\#Parameters.}
The number of parameters corresponds to the total number of learnable weights and biases in a layer. It serves as a proxy for memory footprint and weight-transfer cost during inference.
\noindent\textbf{\#Activations.}
The number of activations is the total number of output elements produced by a layer (computed as \texttt{output.numel()}). It reflects the amount of intermediate data generated and transferred through memory~\cite{sze_efficient_2017}.
\noindent\textbf{Batch Size.}
Batch size determines the number of inputs processed simultaneously. Our experiments show that it significantly affects the energy consumption per input and is therefore included as a feature.
\noindent\textbf{Input Size.}
Input size represents the dimensionality of the input (e.g., image resolution or prompt length). It directly impacts both computation and memory usage and strongly influences energy consumption.

\section{Comparison against SOTA}

\label{app:sota}

\begin{figure}[ht]
    \centering
    \includegraphics[width=0.7\linewidth]{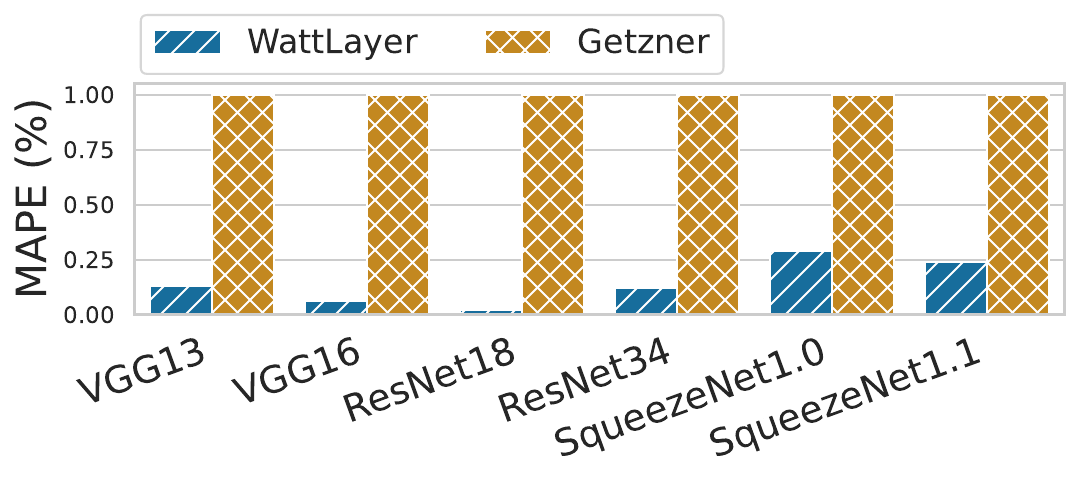}
    \caption{Comparison to SOTA model 'Getzner' \cite{getzner_accuracy_2023}.}
\label{f:sota}
\end{figure}

\end{document}